\documentclass[runningheads]{llncs}

\usepackage{eccv}

\usepackage{eccvabbrv}

\usepackage{graphicx}
\usepackage{booktabs}
\usepackage{subcaption} 

\usepackage[accsupp]{axessibility}  

\usepackage{hyperref}

\usepackage{orcidlink}

\begin{document}

\title{ChartZero: Synthetic Priors Enable Zero Shot Chart Data Extraction} 

\titlerunning{ChartZero}

\author{Md Touhidul Islam, Yasir Mahmud, Sujan Kumar Saha,
Mark Tehranipoor, Farimah Farahmandi}

\authorrunning{M.~T.~Islam et al.}

\institute{University of Florida\\
\email{mdtouhidul.islam@ufl.edu, ymahmud@ufl.edu, sujansaha@ufl.edu\\
tehranipoor@ufl.edu, farimah@ece.ufl.edu}}

\maketitle

\begin{abstract}
Automated data extraction from line charts remains fundamentally bottlenecked by extreme stylistic diversity and a severe scarcity of comprehensively annotated, real-world datasets. Current end-to-end pipelines depend heavily on costly manual annotations, crippling their ability to generalize across arbitrary aesthetics and grid layouts. Furthermore, existing models suffer from two critical failure modes during reconstruction. First, extracting thin, intersecting curves frequently causes structural fragmentation and the erasure of fine visual details, as standard architectures struggle against complex backgrounds. Second, semantic association is notoriously error-prone; current pipelines rely on rigid spatial heuristics that easily break down against the unpredictable legend placements of in-the-wild charts. Finally, measuring true progress is hindered by evaluation protocols that assess isolated sub-tasks rather than holistic, end-to-end data reconstruction. To address these foundational issues, we introduce \textit{ChartZero}, a parsing framework that leverages synthetic priors to enable robust zero-shot chart data extraction. By training exclusively on a purely synthetic dataset of simple mathematical functions, our model completely bypasses the real-world annotation bottleneck. We overcome curve fragmentation via a novel Global Orthogonal Instance (GOI) loss, and replace brittle spatial rules with an open-vocabulary, Vision-Language Model (VLM)-guided legend matching strategy. Accompanied by a new metric and benchmark specifically designed for full end-to-end reconstruction, our evaluations demonstrate that \textit{ChartZero} significantly advances generalized plot digitization without requiring real-world supervision.
\keywords{Line Chart Extraction \and Instance Segmentation \and Synthetic Priors \and Zero-Shot Learning \and Vision Language Models \and Chart Reconstruction}
\end{abstract}

\section{Introduction}
\label{sec:intro}

Automated extraction of data from line charts remains challenging because useful reconstruction requires solving three coupled steps: curve tracing, legend association, and axis-aware numeric mapping. Errors in any one stage can invalidate the final output, so strong intermediate scores do not necessarily translate to reliable end-to-end reconstruction. This coupling makes chart parsing fundamentally different from single-task image understanding problems.

Robustness is further limited by large stylistic variation in ``in-the-wild'' charts, including thin overlapping curves, diverse legend layouts, non-uniform axis formatting, and document artifacts from PDF rasterization or scanning. Most systems rely on expensive real-world annotations and often overfit to specific visual distributions, which weakens transfer to unseen chart styles.

Recent Vision-Language Models (VLMs) can attempt full reconstruction in one pass, but their reliability and numeric precision on dense plots remain under-explored, especially when multiple curves intersect or legends use non-standard encodings. These gaps motivate a data-efficient approach that separates topology extraction from semantic association while still being evaluated with a strict end-to-end criterion.

\begin{figure}[tb]
    \centering
    \begin{subfigure}[b]{\linewidth}
        \centering
        \includegraphics[width=\linewidth]{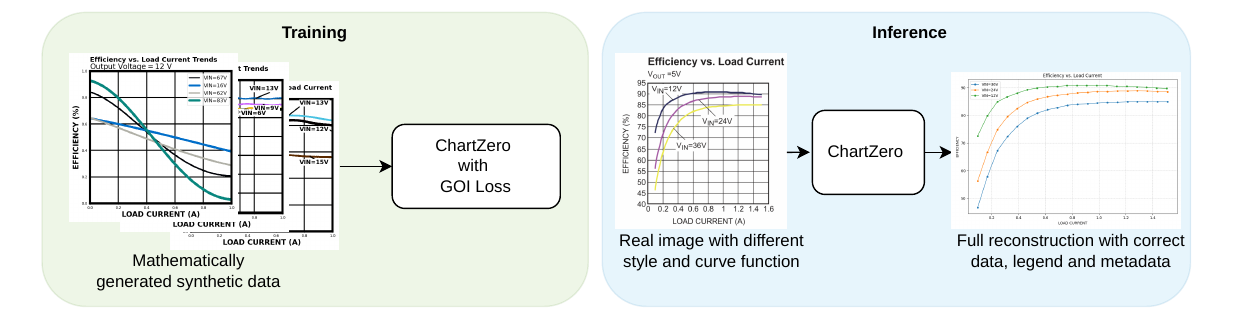}
        \caption{ChartZero achieves zero-shot generalization to real-world charts despite being trained exclusively on synthetic data.}
        \label{fig:flow_training}
    \end{subfigure}
    
    \vspace{1em} 
    
    \begin{subfigure}[b]{\linewidth}
        \centering
        \includegraphics[width=\linewidth]{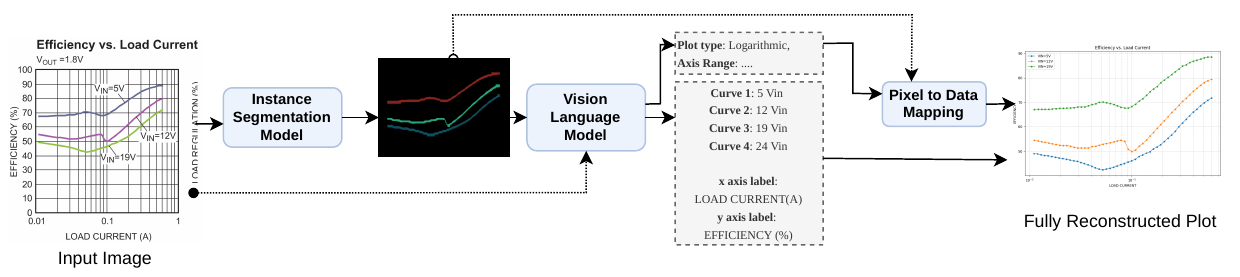}
        \caption{Overall Inference Pipeline}
        \label{fig:flow_inference}
    \end{subfigure}
    
    \caption{Overview of the ChartZero framework. (a) The training pipeline, illustrating how the segmentation network is trained exclusively on synthetic mathematical curves. (b) The end-to-end inference pipeline, detailing the instance segmentation process combined with the open-vocabulary VLM-guided legend matching.}
    \label{fig:chartzero_full_flow}
\end{figure}

We propose \textit{ChartZero}, a data-efficient framework that combines topology-aware instance segmentation with open-vocabulary semantic mapping. The core design principle is to learn geometry from synthetic priors and defer open-ended semantic interpretation to a VLM. ChartZero is trained only on synthetic data yet generalizes to diverse real charts. Our main contributions are:

\begin{itemize}
    \item \textbf{Zero-Shot Instance Segmentation via Synthetic Priors:} A CNN-based segmentor with the proposed GOI loss, trained exclusively on synthetic charts generated from 20 mathematical function families.

    \item \textbf{Open-Vocabulary Legend Mapping:} A masked, VLM-guided legend association strategy that handles diverse legend styles without brittle hand-crafted rules.

    \item \textbf{Comprehensive Evaluation Framework \& Benchmarking:} A full-reconstruction benchmark and ChartRM metric, with extensive comparisons against chart-specific pipelines and state-of-the-art VLMs.
\end{itemize}

\section{Related Works}

The task of automated chart data extraction has been approached from several directions. We categorize the existing literature into three primary domains: task-specific chart extraction models, general-purpose segmentation networks, and Vision-Language Models (VLMs).

\subsection{Task-Specific Chart Extraction}
Historically, task-specific chart extraction models fall into two categories: \textbf{keypoint-based} and \textbf{segmentation-based} approaches.

\textbf{Keypoint-Based Approaches:} These methods detect discrete coordinates along plotted curves and link them to reconstruct data series. ChartOCR (2021) \cite{Luo_2021_WACV} and Kato et al. (2022) \cite{Kato_2022_WACV} rely on grouping and linear programming, while transformer-based models such as LineEX \cite{P._2023_WACV}, LineFormer \cite{lal2023lineformerrethinkinglinechart}, and Efficient Extraction \cite{YANG2025101259} improve localization. Although effective for simple plots, they struggle with complex point associations in densely overlapping charts.

\textbf{Segmentation-Based Approaches:} These methods predict dense pixel masks representing curves. Early work like FigureSeer (2016) \cite{10.1007/978-3-319-46478-7_41} introduced chart segmentation, while newer architectures such as ChartLine (2024) \cite{yang2024chartline} use spatial-sequence fusion to track complex curves. However, segmentation models often fail to preserve distinct instances at line intersections, leading to merge errors.

Despite progress, domain-specific models remain limited by two issues addressed in this work:

\begin{enumerate}
    \item \textbf{Supervision Bottleneck and Data Scarcity:} Most state-of-the-art systems rely on large annotated chart datasets, which are expensive to curate and fail to capture the wide diversity of real-world chart styles, limiting generalization to unseen visual formats.

    \item \textbf{Absence of End-to-End Reconstruction:} Prior work typically targets isolated tasks (e.g., axis detection or curve segmentation) while overlooking full data reconstruction. In particular, \textit{legend mapping}—associating curves with semantic labels—is often missing, leaving extracted values without contextual meaning.
\end{enumerate}

As shown in Table~\ref{tab:dataset_comparison_compact}, unlike pipelines relying on real or mixed supervision, \textit{ChartZero} mitigates these limitations by leveraging synthetic mathematical priors for complete end-to-end reconstruction.

\begin{table}[t]
  \centering
  \caption{Comparison of training data types and evaluation methodologies. Unlike existing methods that rely on real-world data or mixed supervision, ChartZero demonstrates that training exclusively on synthetic mathematical priors is sufficient for full reconstruction tasks.}
  \label{tab:dataset_comparison_compact}
  \begin{tabular}{@{}lcc@{}}
    \toprule
    \textbf{Paper} & \textbf{Training Dataset} & \textbf{Full Recon. Eval} \\
    \midrule
    ChartOCR (2021)\cite{Luo_2021_WACV} & Real & No \\
    Linear Programming (2022)\cite{Kato_2022_WACV} & Real + Synth & No \\
    LineEX (2023) \cite{P._2023_WACV} & Real + Synth & No \\
    LineFormer (2023) \cite{lal2023lineformerrethinkinglinechart} & Real + Synth & No \\
    FigureSeer (2016)\cite{10.1007/978-3-319-46478-7_41} & Real & \textbf{Yes} \\
    ChartLine (2024)\cite{yang2024chartline} & Real + Synth & No \\
    Efficient Extraction (2025)\cite{YANG2025101259} & Real + Synth & No \\
    \midrule
    \textbf{ChartZero (Ours)} & \textbf{Synth} & \textbf{YES} \\
    \bottomrule
  \end{tabular}
\end{table}

\subsection{General-Purpose Segmentation Models}
Beyond chart-specific pipelines, generic segmentation architectures such as FCN \cite{long2015fcn}, U-Net \cite{ronneberger2015unet}, Mask R-CNN \cite{he2017maskrcnn}, and DeepLabv3+ \cite{chen2018deeplabv3plus} established strong dense prediction baselines. More recent promptable foundation models, including SAM \cite{kirillov2023segmentanything} and SAM 3 \cite{carion2025sam3segmentconcepts}, excel at zero-shot segmentation in natural images. However, these models present two critical shortcomings for chart digitization. First, their strictly pixel-level operation cannot semantically associate segmented curves with textual legends. Second, as demonstrated in our evaluations, they consistently fail to maintain the topological integrity of ultra-thin, overlapping lines, resulting in severely fragmented masks.

\subsection{Vision-Language Models (VLMs)}
Recent advancements in multimodal architectures have introduced Vision-Language Models capable of sophisticated visual reasoning and chart data extraction. Foundational cross-modal pretraining with CLIP \cite{radford2021clip} and instruction-tuned multimodal models such as LLaVA \cite{liu2023llava}, Qwen-VL \cite{bai2023qwenvl}, GPT-4 \cite{openai2023gpt4}, and Gemini \cite{team2023gemini} have accelerated chart-level reasoning. Models such as GPT-4o \cite{openai2023gpt4}, Gemini-family models \cite{team2023gemini}, Claude \cite{anthropic2024claude3}, and Gemma \cite{team2024gemma} can generate full reconstructions from chart images, including legend mapping, axis label identification, and plot title extraction. Currently, these large VLMs stand as the primary alternative capable of performing end-to-end reconstruction. However, they often struggle with precision mapping in highly cluttered plots.

\section{Methodology}
\subsection{Synthetic Dataset Design}

\begin{figure}[t]
    \centering
    \includegraphics[width=0.32\linewidth]{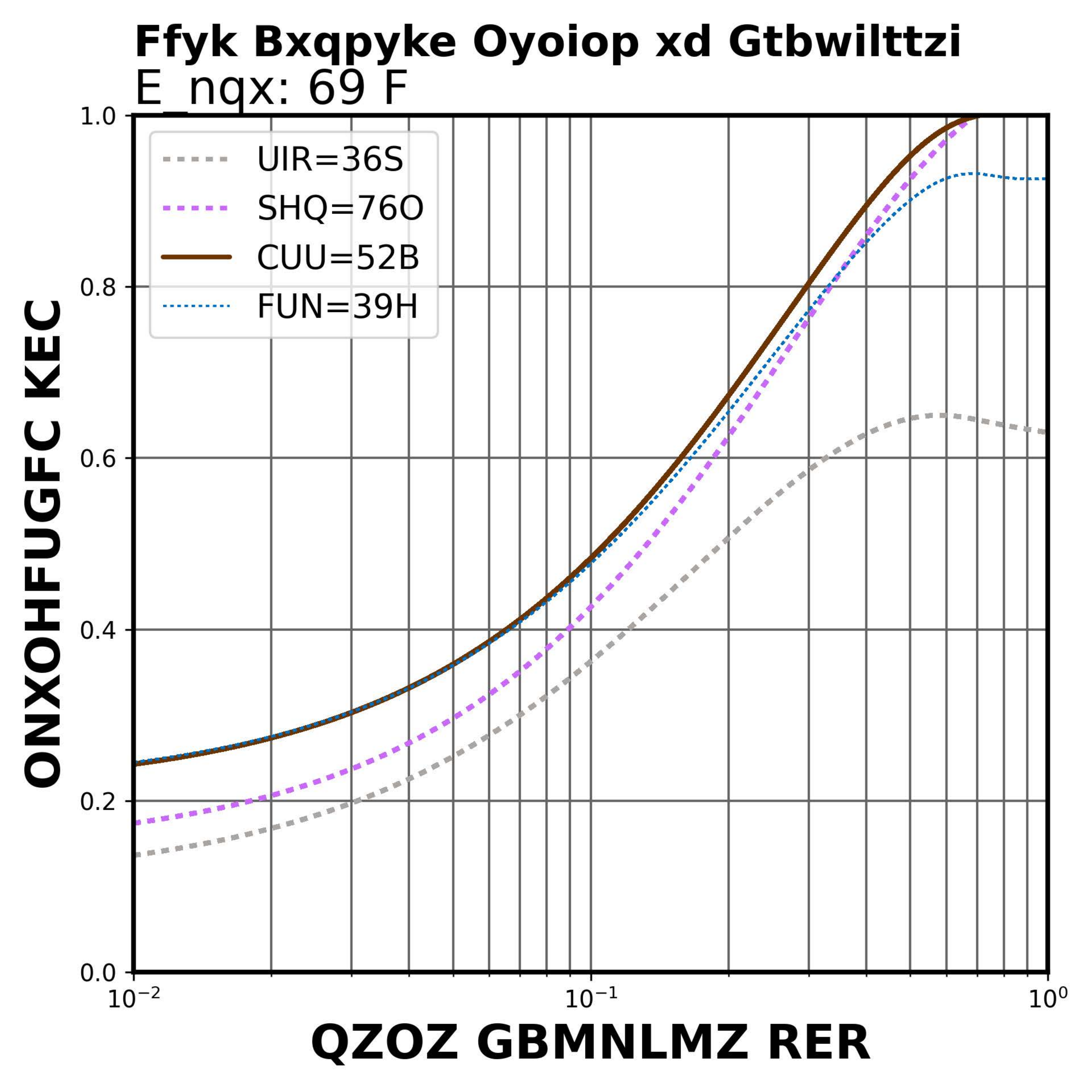}\hfill
    \includegraphics[width=0.32\linewidth]{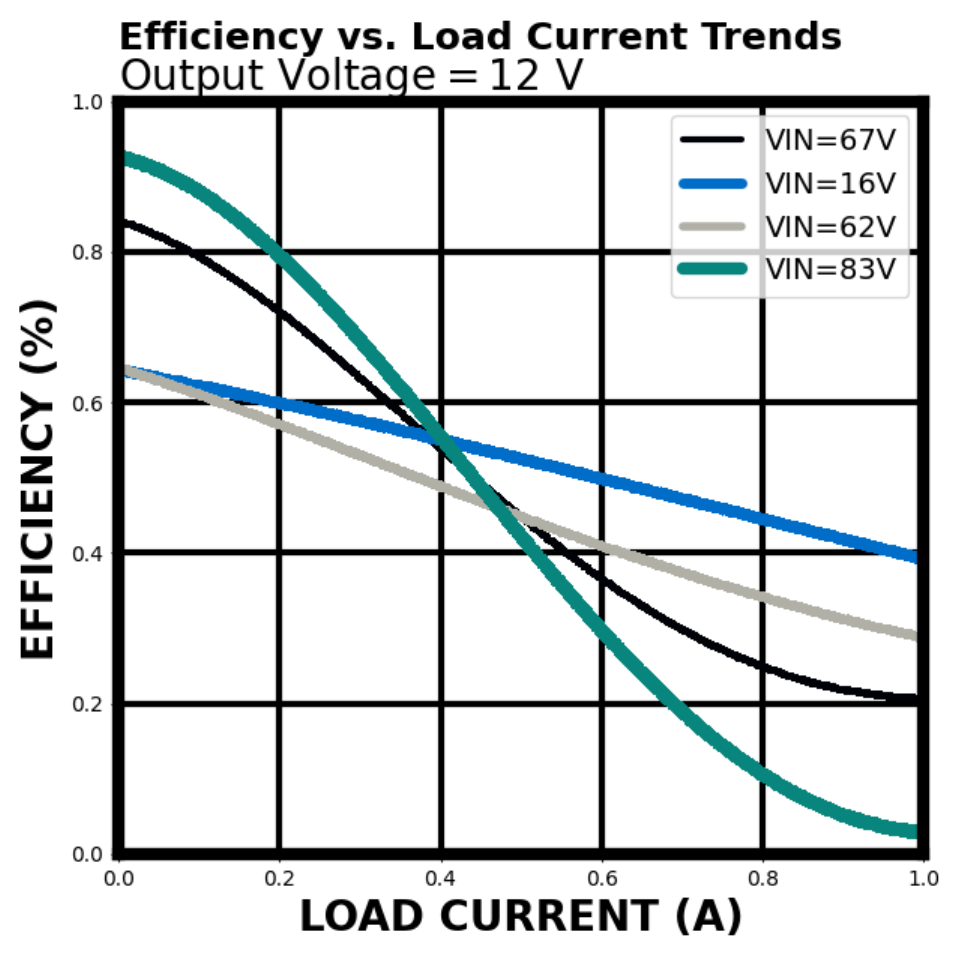}\hfill
    \includegraphics[width=0.32\linewidth]{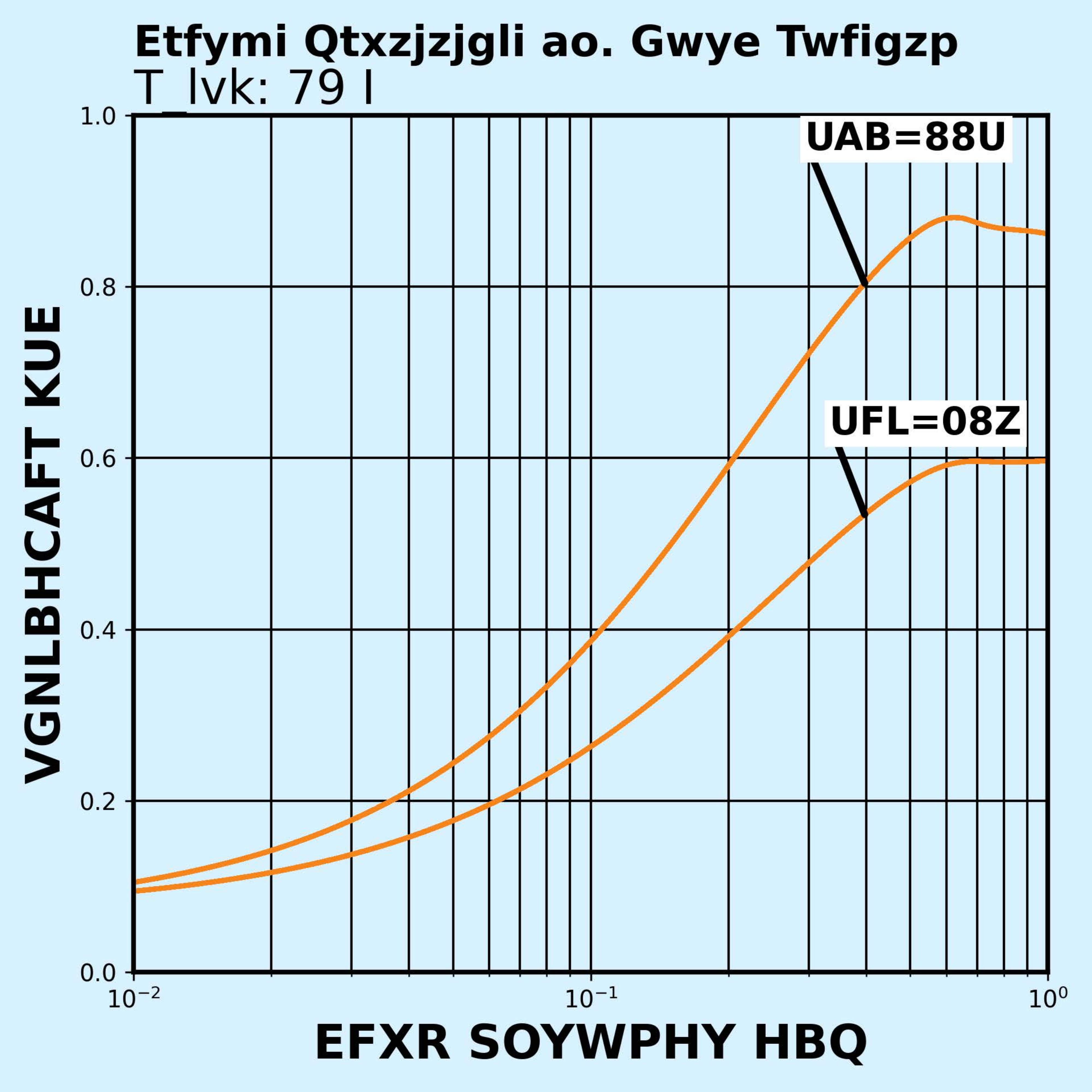}

    \caption{Samples from the synthetic dataset featuring randomized visual and structural parameters.}
    \label{fig:synthetic_samples}
\end{figure}

Real-world charts vary widely in style, but most of this variation is irrelevant to curve geometry. \textit{ChartZero} decouples low-level topological extraction from high-level semantic reasoning (handled by the VLM), so the segmentor focuses on geometric continuity under overlap.

We build a synthetic training set of 100,000 charts in \texttt{matplotlib}, generated from 20 parameterized function families (sinusoidal, logistic, power-law, damped, and exponential). This basis captures common geometric regimes while remaining balanced for stable training.

During rendering, we randomize curve parameters, overlap patterns, line styles (solid/dashed/dotted), colors, axis labels, titles, legend styles, grids, and background textures, with additional data-level noise.

We further apply online augmentations (scaling, translation, mild affine distortion, color jitter, Gaussian blur, and JPEG compression) to improve robustness to scan and compression artifacts. Ground-truth instance masks are obtained directly from the renderer, and we use an 80/20 train/validation split. Figure \ref{fig:synthetic_samples} shows some images from the synthetic dataset.

\subsection{Instance Segmentation Network}
The core of our instance segmentation framework is built upon a dual-headed U-Net architecture. To explicitly provide the network with spatial awareness---a crucial inductive bias for tracing long, continuous curves across a chart---we augment the standard U-Net backbone with a CoordConv layer. This layer concatenates normalized Cartesian coordinates ($x, y \in [-1, 1]$) directly to the input image channels before feature extraction. Furthermore, to ensure stable training on high-resolution chart images with small batch sizes, we replace standard Batch Normalization with Group Normalization across all convolutional blocks.

The network bifurcates into two distinct spatial prediction heads at the final decoding stage:
\begin{enumerate}
    \item \textbf{Semantic Head:} A $1 \times 1$ convolutional layer predicting a dense, binary semantic mask that separates all foreground curve pixels from the background.
    \item \textbf{Instance Embedding Head:} A parallel $1 \times 1$ convolutional layer that outputs a highly dimensional, dense feature embedding for every pixel. To strictly align with our topology-aware graph loss formulation, the output of this head is explicitly $L_2$-normalized, projecting every pixel's discriminative feature vector onto a continuous unit hypersphere.
\end{enumerate}
This dual-head design allows the network to simultaneously perform highly local foreground detection while independently learning the discriminative metric space necessary to decouple overlapping and intersecting line instances.

\subsection{Global Orthogonal Instance Loss}
\label{sec:loss_function}

\begin{figure}[t]
    \centering
    \begin{subfigure}[b]{0.48\textwidth}
        \centering
        \includegraphics[width=0.7\linewidth]{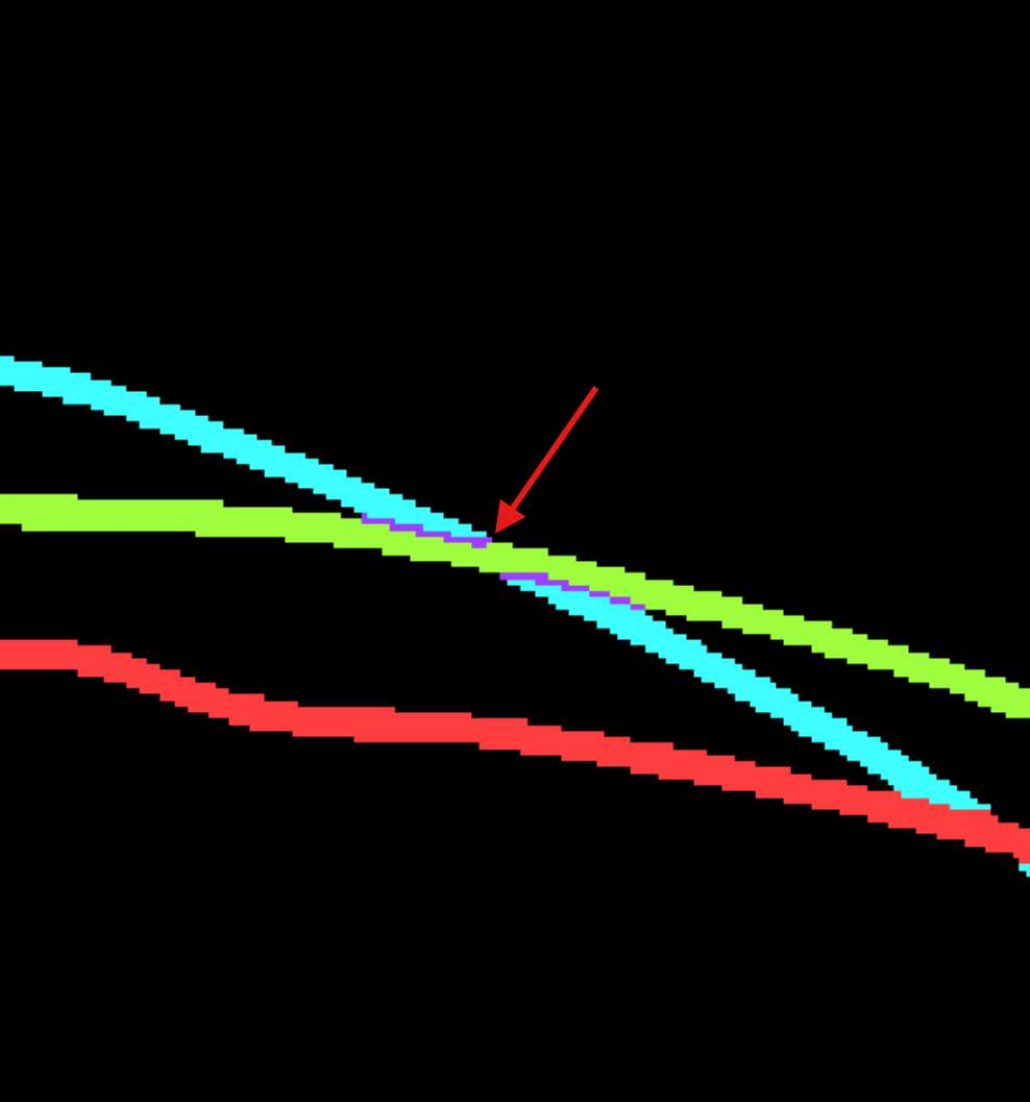}
        \caption{Baseline Failure Case at Intersection}
        \label{fig:intersection}
    \end{subfigure}
    \hfill
    \begin{subfigure}[b]{0.48\textwidth}
        \centering
        \includegraphics[width=\linewidth]{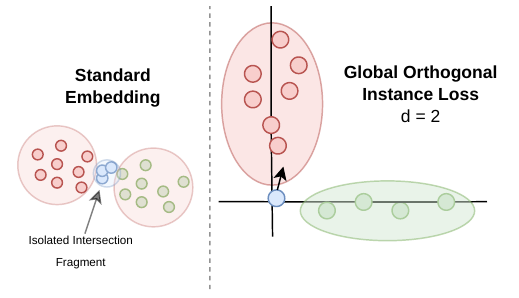}
        \caption{Our GOI Loss}
        \label{fig:goi_visual}
    \end{subfigure}
    \caption{Visualizing the impact of the Global Orthogonal Instance (GOI) loss. (a) Standard clustering methods frequently get confused where thin curves intersect, leading to broken and fragmented lines. (b) Our GOI loss preserves topological continuity and handles line overlaps robustly.}
    \label{fig:goi_loss_comparison}
\end{figure}

Accurate instance segmentation of thin, curvilinear structures such as lines in plot images presents a unique challenge: lines frequently intersect, leading to severe fragmentation when assigning instance identities. Traditional discriminative loss formulations pull pixels toward instance centroids and push centroids apart, but they fail to robustly handle the topological ambiguity at intersections, often isolating minimal intersection regions into spurious, independent instances.

To address this, we introduce the Global Orthogonal Instance (GOI) Loss. This formulation projects pixel embeddings onto a unit hypersphere, forcing distinct line instances to occupy mutually orthogonal subspaces, while explicitly penalizing the fragmentation of small intersection clusters. Figure \ref{fig:goi_visual} illustrates the core geometric intuition behind our proposed GOI loss.

\textbf{Hypersphere Projection and Centroid Computation.} 
Let $\mathbf{E} \in \mathbb{R}^{D \times H \times W}$ represent the dense $D$-dimensional pixel embeddings predicted by the network. We first project these embeddings onto the unit hypersphere by applying $L_2$ normalization along the channel dimension, yielding $\mathbf{e}_i = \mathbf{E}_i / ||\mathbf{E}_i||_2$ for each pixel $i$.
For a given image containing $N$ active ground-truth instances, let $\mathcal{S}_k$ denote the set of pixels belonging to instance $k$, and $M_k = |\mathcal{S}_k|$ be its pixel count. The centroid $\mathbf{c}_k$ of instance $k$ is computed as the $L_2$-normalized mean of its constituent pixel embeddings:
\begin{equation}
    \mathbf{c}_k = \frac{\boldsymbol{\mu}_k}{||\boldsymbol{\mu}_k||_2}, \quad \text{where} \quad \boldsymbol{\mu}_k = \frac{1}{M_k} \sum_{i \in \mathcal{S}_k} \mathbf{e}_i
\end{equation}

\textbf{Pull Loss (Intra-Instance Cohesion).} 
To ensure that pixels belonging to the same line segment form a tight cluster, we maximize the cosine similarity between each pixel embedding $\mathbf{e}_i$ and its assigned instance centroid $\mathbf{c}_k$. The pull loss is defined as:
\begin{equation}
    \mathcal{L}_{\text{pull}} = \frac{1}{N} \sum_{k=1}^{N} \frac{1}{M_k} \sum_{i \in \mathcal{S}_k} \left( 1 - \mathbf{c}_k^\top \mathbf{e}_i \right)
\end{equation}
Because both the embeddings and the centroids lie on the unit hypersphere, their dot product directly computes the cosine similarity.

\textbf{Push Loss (Inter-Instance Orthogonality).} 
To separate distinct line instances, we enforce orthogonality between their centroids. Let $\mathcal{V}$ denote the subset of ``large'' instances (e.g., $M_k \ge \tau$, where $\tau$ is a spatial threshold representing the minimum size of an independent line). We compute the pairwise cosine similarity between all valid distinct centroids $\mathbf{c}_u, \mathbf{c}_v \in \mathcal{V}$ and penalize their squared magnitude:
\begin{equation}
    \mathcal{L}_{\text{push}} = \frac{1}{|\mathcal{V}|(|\mathcal{V}|-1)} \sum_{u \in \mathcal{V}} \sum_{\substack{v \in \mathcal{V} \\ v \neq u}} \left( \mathbf{c}_u^\top \mathbf{c}_v \right)^2
\end{equation}
Squaring the dot product ensures that centroids are pushed toward orthogonal directions ($\mathbf{c}_u^\top \mathbf{c}_v \to 0$) globally across the hypersphere, maximizing the inter-class angular margin.

\textbf{Merge Loss (Small-Cluster Penalization).} 
At graph intersections, the network may incorrectly fragment a contiguous line by assigning a unique embedding to the small overlapping region. To regularize this, we explicitly penalize the isolation of ``small'' instances $\mathcal{U}$ (where $M_k < \tau$). Rather than pushing a small intersection cluster away from the main lines, we force it to assimilate into the subspace of at least one adjacent large line.
For each small instance $s \in \mathcal{U}$, we maximize its centroid's similarity to the nearest large instance centroid $l \in \mathcal{V}$:
\begin{equation}
    \mathcal{L}_{\text{merge}} = \frac{1}{|\mathcal{U}|} \sum_{s \in \mathcal{U}} \left( 1 - \max_{l \in \mathcal{V}} (\mathbf{c}_s^\top \mathbf{c}_l) \right)
\end{equation}
This term effectively eliminates intersection fragmentation by heavily penalizing small, highly dissimilar sub-clusters.

\textbf{Total Objective.} 
The final GOI loss is computed as a weighted sum of the cohesion, orthogonality, and intersection penalization terms:
\begin{equation}
    \mathcal{L}_{\text{GOI}} = \alpha \mathcal{L}_{\text{pull}} + \beta \mathcal{L}_{\text{push}} + \gamma \mathcal{L}_{\text{merge}}
\end{equation}
where $\alpha, \beta$, and $\gamma$ are hyperparameters governing the relative importance of intra-class variance, inter-class separation, and topology preservation, respectively.

\subsection{Legend Mapping}
\begin{figure}[t]
    \centering
    \begin{subfigure}[t]{0.32\linewidth}
        \centering
        \includegraphics[width=\linewidth,height=0.15\textheight,keepaspectratio]{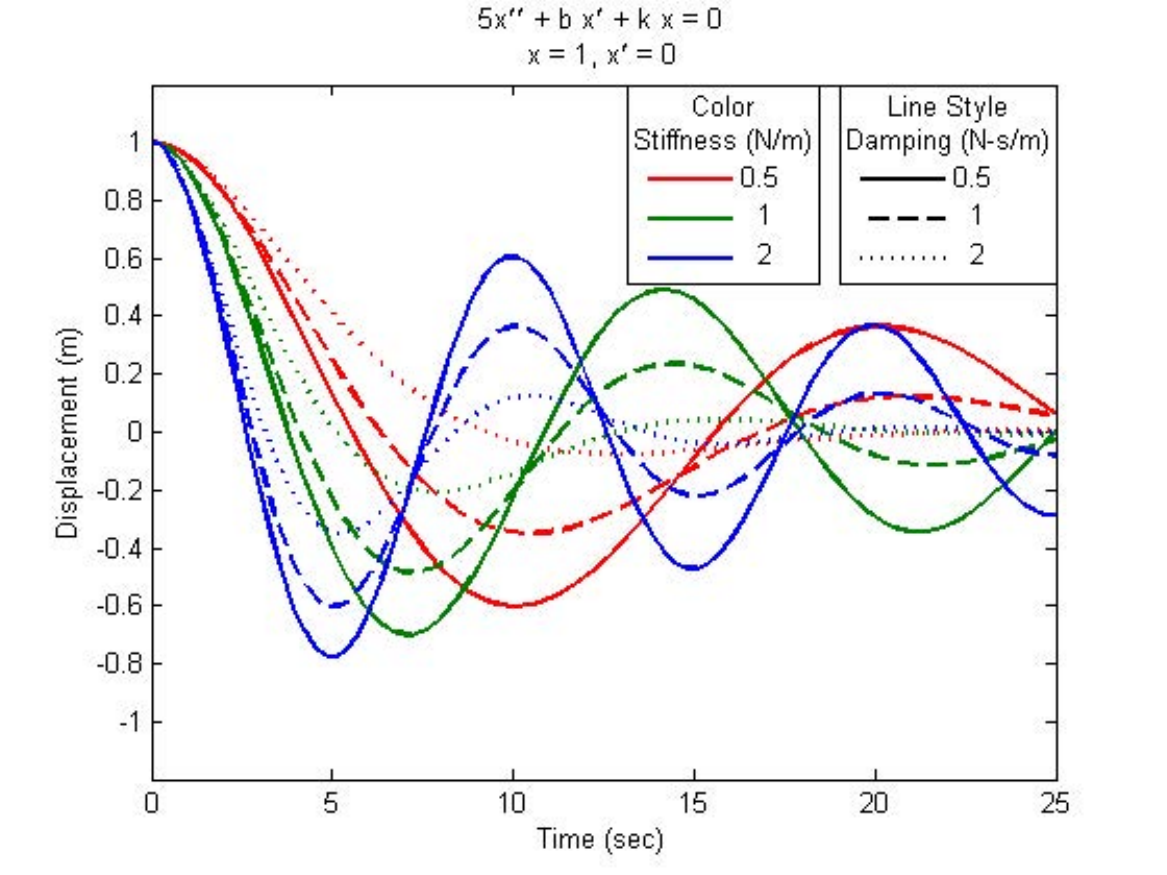}
        \caption{Composite legend with separate encodings for color and line style.}
        \label{fig:legend-type-composite}
    \end{subfigure}\hfill
    \begin{subfigure}[t]{0.32\linewidth}
        \centering
        \includegraphics[width=\linewidth,height=0.18\textheight,keepaspectratio]{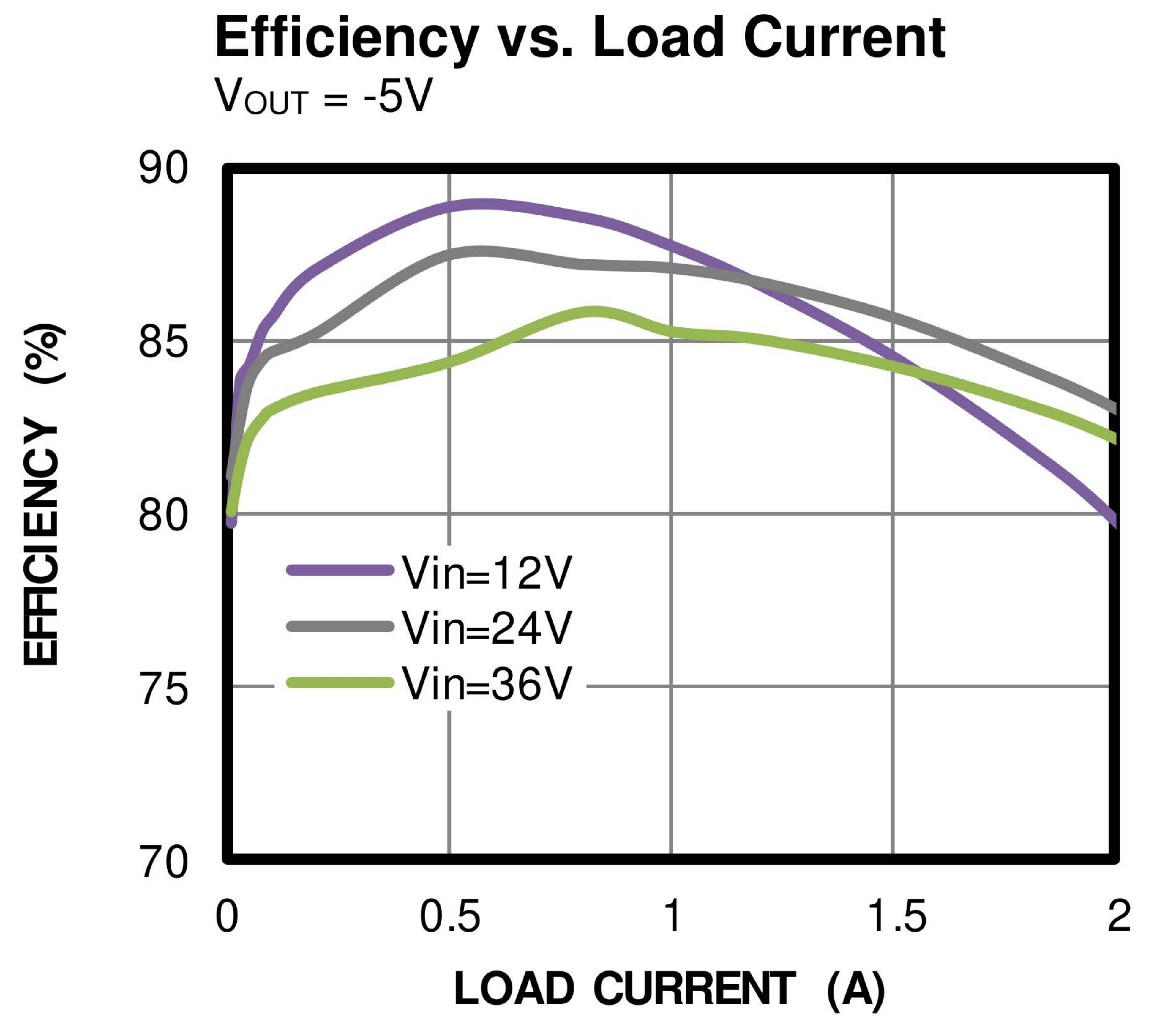}
        \caption{In-plot legend block overlaying the chart area.}
        \label{fig:legend-type-inline}
    \end{subfigure}\hfill
    \begin{subfigure}[t]{0.32\linewidth}
        \centering
        \includegraphics[width=\linewidth,height=0.18\textheight,keepaspectratio]{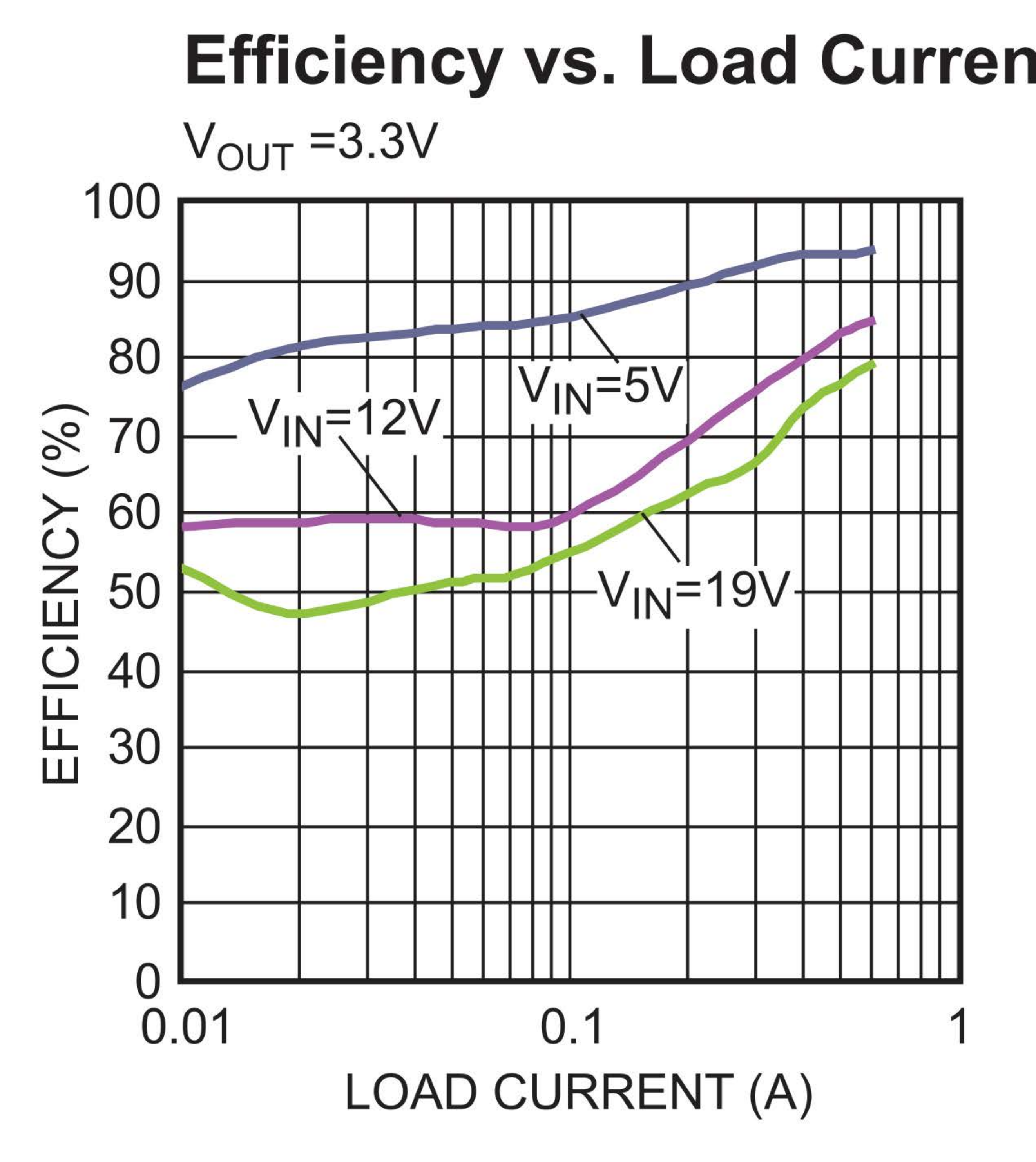}
        \caption{Direct line labeling with leader arrows on a dense grid.}
        \label{fig:legend-type-direct}
    \end{subfigure}
    \caption{Varieties of legend presentation styles used to evaluate open-vocabulary legend mapping.}
    \label{fig:legend-style-varieties}
\end{figure}

Legend mapping is difficult because legend layouts and styles vary widely (position, structure, color, marker, and line-style combinations). Figure~\ref{fig:legend-style-varieties} shows representative formats, including composite keys, in-plot legend blocks, and direct annotations.

We use an open-vocabulary, VLM-guided strategy: for each predicted curve instance, we create a query image that keeps only the target curve visible and masks the others, then prompt the VLM to return the corresponding legend label. This avoids brittle rule-based matching (e.g., explicit legend box detection and style tracing) and enables robust zero-shot mapping across diverse legend designs.
In our implementation, this semantic module uses Gemma 3 27B \cite{team2024gemma} as the underlying VLM.

\subsection{End-to-End Data Extraction}
The final objective of the pipeline is to map the semantic, pixel-level curve segments into their corresponding numerical data arrays. First, we localize the spatial bounding box of the plot area using an OpenCV-based contour approximation pipeline (Canny edge detection followed by \texttt{approxPolyDP}), isolating the primary quadrangle. Simultaneously, we prompt the VLM with the original image to extract crucial quantitative metadata, including the axes labels, the discrete numerical ranges of the X and Y axes, and their respective mathematical scales (linear or logarithmic). By using the localized pixel boundaries and the VLM-extracted numerical ranges, we construct an affine transformation matrix. This matrix directly maps the segmented curve pixel coordinates into real-world data points, successfully outputting a fully reconstructed, machine-readable data table representing the original chart.

\section{Results}
\subsection{The ChartZero Benchmark}
We introduce the \textit{ChartZero Benchmark} to evaluate the setting that matters in practice: \emph{full chart reconstruction}, not isolated sub-tasks. Existing resources such as FigureSeer \cite{10.1007/978-3-319-46478-7_41}, ExcelChart400K \cite{Luo_2021_WACV}, Adobe CHART-Synthetic \cite{adobe_chart_synthetic2019}, and LineEX \cite{P._2023_WACV} are valuable, but they do not consistently provide all labels required to verify end-to-end correctness (trace geometry + legend binding + axis semantics + scale metadata). This makes it difficult to compare methods on real downstream utility.

To close this gap, we construct a 1,000-chart benchmark with complete labels per chart: per-series trace masks, legend-to-curve associations, axis names, numeric ranges, and linear/log scale metadata. The construction pipeline is reproducible: (i) PDF corpus mining from heterogeneous technical and business documents, (ii) YOLO-based plot-region extraction, (iii) line-chart filtering, (iv) dense annotation in CVAT \cite{cvat2026}, and (v) adjudication-driven quality control.

Label quality is enforced with a two-pass protocol. Each chart is independently annotated by two annotators; disagreements are resolved by a third-pass adjudicator using fixed guidelines. We further run consistency checks by projecting trace masks back to chart coordinates and validating axis/legend fields against the rendered text. This produces high annotation reliability (legend mapping $\kappa=0.96$, axis metadata exact-match $98.7\%$, and median trace-boundary error $<1.0$ pixel), necessary for strict evaluation with ChartRM.

The benchmark emphasizes diversity rather than convenience. It includes sparse and dense multi-series charts, multiple legend presentation styles, mixed axis-scale configurations, and common PDF artifacts (compression, scanning noise, and low-contrast overlays). Table~\ref{tab:benchmark_profile} summarizes this diversity profile. Additional construction, annotation protocol, and quality-control details are provided in the supplementary material.

\begin{table}[t]
    \centering
    \caption{Diversity profile of the ChartZero Benchmark (1,000 charts).}
    \label{tab:benchmark_profile}
    \begin{tabular}{lp{7.6cm}}
        \toprule
        \textbf{Diversity Aspect} & \textbf{Distribution} \\
        \midrule
        Series density & 1--2 series: 286; 3--4 series: 451; 5+ series: 263 (max 11). \\
        Legend presentation style & External legend box: 397; in-plot legend: 241; direct labeling: 209; composite encodings: 153. \\
        Axis-scale configuration & Linear--linear: 612; linear--log: 173; log--linear: 126; log--log: 89. \\
        Document artifact regime & Clean digital: 458; compression/noise: 292; scanned/print artifacts: 157; low-contrast overlap: 93. \\
        \bottomrule
    \end{tabular}
\end{table}

\subsection{Evaluation Metric: ChartRM}

We introduce the Chart Reconstruction Metric (ChartRM) to evaluate the true, practical utility of the extracted data. 

The philosophy behind this metric is based on a strict hierarchy of data utility:
\begin{enumerate}
    \item \textbf{Data Accuracy ($\Gamma_{data}$):} First, we must accurately trace the data lines. A score of 1 means 100\% correct data capture, and 0 means 0\% correct.
    \item \textbf{Legend Matching ($\Psi_{legend}$):} However, perfectly extracted data is completely useless if it is attached to the wrong legend.
    \item \textbf{Axis Labels ($\Phi_{axis}$):} Finally, even if we perfectly extract the data and match the legend, it is all useless if we get the axis labels wrong. If we do not know what the axes represent, the entire plot's data has no meaning. 
\end{enumerate}

Because an incorrect axis label invalidates all the data inside the plot, the axis score acts as a gatekeeper and multiplies everything else from the outside. The metric is defined as:

$$\text{ChartRM} = \Phi_{axis} \times \left( \frac{1}{N} \sum_{i=1}^{N} \Psi_{legend, i} \cdot \Gamma_{data, i} \right)$$

where $N$ is the total number of lines (data series) in the chart. The three components work as follows:

\begin{itemize}
    \item \textbf{Axis Label Accuracy ($\Phi_{axis} \in \{0, 1\}$):} A simple pass or fail check. If the model incorrectly identifies the axis labels, this value is 0, which brings the score for the entire chart down to 0.

    \item \textbf{Legend Matching ($\Psi_{legend, i} \in \{0, 1\}$):} A pass or fail check for each specific line $i$. If the model links the line to the wrong legend, its score becomes 0, throwing out that specific line's data.

    \item \textbf{Data Accuracy ($\Gamma_{data, i} \in [0, 1]$):} For lines that are matched to the correct legend, this measures how close the predicted line $\mathbf{\hat{y}}_i$ is to the real data $\mathbf{y}_i$. We calculate it using the following equation:
    $$\Gamma_{data, i} = \exp\left( -\lambda \cdot \frac{\text{RMSE}(\mathbf{y}_i, \mathbf{\hat{y}}_i)}{\text{Range}(Y)} \right)$$
    This formula gives a score close to 1 for highly accurate traces and penalizes lines that drift away from the real data.
\end{itemize}

\subsection{Evaluating Data Extraction}

Table~\ref{tab:compare_data_extraction} compares \textit{ChartZero} against the general-purpose segmentor SAM 3 and chart-specific architectures (LineFormer, LineEX) across three datasets: our \textit{ChartZero Benchmark}, the \textit{ExcelData400k} test set, and the \textit{LineEX} test set. The baseline models exhibit severe performance degradation on our benchmark, demonstrating limited generalization beyond their training distributions. For instance, while SAM 3 shows strong zero-shot capabilities on natural images, it drops to 0.28 IoU on thin, intersecting chart curves, compared to the 0.82 IoU achieved by \textit{ChartZero}.

For numeric reconstruction (NRMSE), this generalization gap is even more pronounced. Chart-specific models struggle to adapt to new data distributions, with LineFormer and LineEX falling to 0.154 and 0.436 NRMSE on our benchmark. In contrast, \textit{ChartZero} establishes a new state-of-the-art on our benchmark (0.028 NRMSE) while remaining highly competitive on the baselines’ own test sets (\textit{ExcelData400k} and \textit{LineEX}). Although trained exclusively on synthetic data and never exposed to the real-world training sets of these benchmarks, these results highlight its strong zero-shot generalization and stylistic robustness.

\begin{table}[htbp]
    \centering
    \caption{Quantitative comparison of instance segmentation performance. ChartZero shows improvements over general-purpose segmentors and achieves competitive performance with chart-specific architectures across three benchmark datasets.}
    \label{tab:compare_data_extraction}
    \resizebox{\textwidth}{!}{
    \begin{tabular}{lcccccc}
        \toprule
        & \multicolumn{2}{c}{\textbf{ChartZero Bench. (Ours)}} & \multicolumn{2}{c}{\textbf{ExcelData400k Testset}} & \multicolumn{2}{c}{\textbf{LineEX Testset}} \\
        \cmidrule(lr){2-3} \cmidrule(lr){4-5} \cmidrule(lr){6-7}
        \textbf{Method} & \textbf{IoU $\uparrow$} & \textbf{NRMSE $\downarrow$}& \textbf{IoU $\uparrow$} & \textbf{NRMSE $\downarrow$}& \textbf{IoU $\uparrow$} & \textbf{NRMSE $\downarrow$}\\
        \midrule
        SAM 3                   & 0.28 & - & 0.23 & - & 0.42 & -\\
        LineFormer              & 0.68 & - & 0.58 & - & 0.64 & -\\
        LineEX                  & - & 0.436 & - & 0.372 & - & 0.121\\
        \midrule
        \textbf{ChartZero (Ours)} & \textbf{0.82} & \textbf{0.028} & \textbf{0.62} & \textbf{0.162} & \textbf{0.69} & \textbf{0.142}\\
        \bottomrule
    \end{tabular}
    }
\end{table}

\subsection{Evaluating legend mapping and axis label extraction}
Table~\ref{tab:legend-axis-result} evaluates semantic parsing quality: legend-line matching (F1) and axis label extraction (Accuracy). We compare our masked VLM strategy against general-purpose VLM baselines (Gemma 27B \cite{team2024gemma}, DeepSeek-VL2 \cite{deepseek2024vl2}, GPT-4o \cite{openai2023gpt4}, Gemini 3 Pro \cite{team2023gemini}).

\textit{ChartZero} achieves the highest scores on both datasets (ChartZero Benchmark: F1 $0.945$, Accuracy $0.962$; LineEX: F1 $0.938$, Accuracy $0.955$), improving over strong baselines such as Gemini 3 Pro (ChartZero Benchmark: $0.892/0.915$). The masked-query design reduces ambiguity in dense charts by isolating one target curve at a time before semantic matching.

\begin{table}[htbp]
    \centering
    \caption{Performance comparison on intermediate semantic tasks across different datasets. ChartZero demonstrates improved results in both legend mapping (F1-score) and axis label extraction (Accuracy) compared to general-purpose Vision Language Models.}
    \label{tab:legend-axis-result}
    \begin{tabular}{lcccc}
        \toprule
        & \multicolumn{2}{c}{\textbf{ChartZero Bench. (Ours)}} & \multicolumn{2}{c}{\textbf{LineEX Testset}} \\
        \cmidrule(lr){2-3} \cmidrule(lr){4-5}
        \textbf{Method} & \textbf{Legend F1 $\uparrow$} & \textbf{Axis Accuracy $\uparrow$} & \textbf{Legend F1 $\uparrow$} & \textbf{Axis Accuracy $\uparrow$} \\
        \midrule
        Gemma 27B               & 0.753 & 0.781 & 0.742 & 0.768 \\
        DeepSeek                & 0.812 & 0.834 & 0.805 & 0.822 \\
        GPT-4o                  & 0.864 & 0.887 & 0.851 & 0.873 \\
        Gemini 3 Pro            & 0.892 & 0.915 & 0.875 & 0.902 \\
        \midrule
        \textbf{ChartZero (Ours)} & \textbf{0.945} & \textbf{0.962} & \textbf{0.938} & \textbf{0.955} \\
        \bottomrule
    \end{tabular}
\end{table}

\subsection{Evaluating full reconstruction}
Table~\ref{tab:full_comparison} reports end-to-end performance using ChartRM, which jointly evaluates curve accuracy, legend association, and axis understanding. Traditional chart extraction pipelines perform poorly under the holistic ChartRM metric (ChartRM $< 0.07$), showing that strong curve tracing alone is insufficient for full reconstruction.

Large VLMs provide stronger end-to-end baselines (e.g., Gemini 3 Pro \cite{team2023gemini}: $0.8542$, GPT 5.2 \cite{openai2023gpt4}: $0.8122$), but still show errors in dense layouts. \textit{ChartZero} achieves $0.9210$ by combining topology-focused curve extraction with masked VLM-based semantic matching, yielding the strongest full-pipeline result in this comparison.

\begin{table}[htbp]
    \centering
    \caption{Full reconstruction performance comparison. ChartZero achieves the highest ChartRM score against both chart-specific methods and large Vision Language Models.}
    \label{tab:full_comparison}
    \begin{tabular}{lc}
        \toprule
        \textbf{Model} & \textbf{ChartRM $\uparrow$} \\
        \midrule
        \multicolumn{2}{l}{\textit{Chart-Specific Methods}} \\
        \midrule
        ChartOCR                & 0.0621 \\
        LineEX                  & 0.0342 \\
        AI Chart Parser         & 0.0148 \\
        \midrule
        \multicolumn{2}{l}{\textit{Vision Language Models}} \\
        \midrule
        GPT-4o                  & 0.4679 \\
        GPT 5.2                 & 0.8122 \\
        Gemini 3 Pro            & 0.8542 \\
        Gemini 2.5              & 0.7243 \\
        Gemma3 27B               & 0.3178 \\
        DeepSeek-VL2                & 0.2748 \\
        Qwen 3.5 122B           & 0.5129 \\
        \midrule
        \textbf{ChartZero (Ours)} & \textbf{0.9210} \\
        \bottomrule
    \end{tabular}
\end{table}

\subsection{Inference Efficiency and Cost Analysis}
We evaluate the deployment trade-off between accuracy, latency, and cost. Unlike single-pass VLMs, \textit{ChartZero} makes $1$ VLM call for axis metadata and $N$ calls for legend mapping, where $N$ is the number of detected curves.

\begin{table}[t]
    \centering
    \caption{Efficiency--accuracy trade-off. Latency and cost are reported under a point-mass curve-count assumption, $P(N{=}4)=1$ (i.e., exactly 4 curves per chart), using current commercial VLM API pricing.}
    \label{tab:efficiency_analysis}
    \begin{tabular}{lcccc}
        \toprule
        \textbf{Method} & \textbf{ChartRM $\uparrow$} & \textbf{Latency (s) $\downarrow$} & \textbf{Cost (\$) $\downarrow$} & \textbf{Scalability} \\
        \midrule
        GPT-4o (End-to-End) & 0.468 & \textbf{4.2} & \textbf{0.01} & High \\
        Gemini 3 Pro (E2E)  & 0.854 & 5.5 & 0.02 & High \\
        Human Annotation    & $\sim$0.99 & 300+ & 5.00+ & Low \\
        \midrule
        \textbf{ChartZero (Ours)} & \textbf{0.921} & 16.8 & 0.05 & Medium \\
        \bottomrule
    \end{tabular}
\end{table}

Table~\ref{tab:efficiency_analysis} shows that \textit{ChartZero} averages 16.8 s and \$0.05 per chart under $P(N{=}4)=1$, versus GPT-4o \cite{openai2023gpt4} at 4.2 s and \$0.01. The additional latency and cost improve reliability (ChartRM $0.921$ vs. $<0.86$ for end-to-end VLMs) while remaining much cheaper and faster than manual transcription. Since our pipeline issues one legend-mapping call per curve, latency and API cost scale approximately linearly with $N$.

\subsection{Ablation Study}
We perform ablations on the ChartZero Benchmark (1,000 charts), averaged over three random seeds. Starting from a U-Net baseline with discriminative loss \cite{Brabandere_2017_CVPR_Workshops}, we incrementally add CoordConv, GOI, and masked VLM. Without masked VLM, legend mapping is unavailable, so ChartRM is not applicable. Table \ref{tab:ablation} shows the ablation result showing strong empirical impact of our new loss function. We use disciriminative loss \cite{Brabandere_2017_CVPR_Workshops} as baseline for embedding based instance segmentation.

\begin{table}[htbp]
    \centering
    \caption{Component-wise ablation on the ChartZero Benchmark validation split. GOI primarily improves geometric fidelity (IoU/NRMSE). ChartRM is reported only when masked VLM legend mapping is enabled.}
    \label{tab:ablation}
    \resizebox{\linewidth}{!}{
    \begin{tabular}{cccc|ccc}
        \toprule
        \textbf{\shortstack{Base\\(U-Net+DL)}} & \textbf{CoordConv} & \textbf{GOI} & \textbf{\shortstack{Masked\\VLM}} & \textbf{IoU $\uparrow$} & \textbf{NRMSE $\downarrow$} & \textbf{ChartRM $\uparrow$} \\
        \midrule
        \checkmark &  &  &  & 0.71 & 0.087 & N/A \\
        \checkmark & \checkmark &  &  & 0.75 & 0.071 & N/A \\
        \checkmark & \checkmark & \checkmark &  & 0.82 & 0.028 & N/A \\
        \checkmark & \checkmark & \checkmark & \checkmark & \textbf{0.82} & \textbf{0.028} & \textbf{0.921} \\
        \bottomrule
    \end{tabular}
    }
\end{table}

CoordConv improves geometric accuracy (NRMSE: $0.087 \rightarrow 0.071$), and GOI provides the largest gain (IoU: $0.75 \rightarrow 0.82$, NRMSE: $0.071 \rightarrow 0.028$). Adding masked VLM does not change IoU/NRMSE but enables end-to-end evaluation, yielding ChartRM $0.921$. Seed variance remains low ($\sigma_{\text{NRMSE}} \leq 0.004$).

\subsection{Failure cases}
\begin{figure}[t]
    \centering
    \begin{subfigure}[b]{0.5\linewidth}
        \centering
        \includegraphics[width=\linewidth]{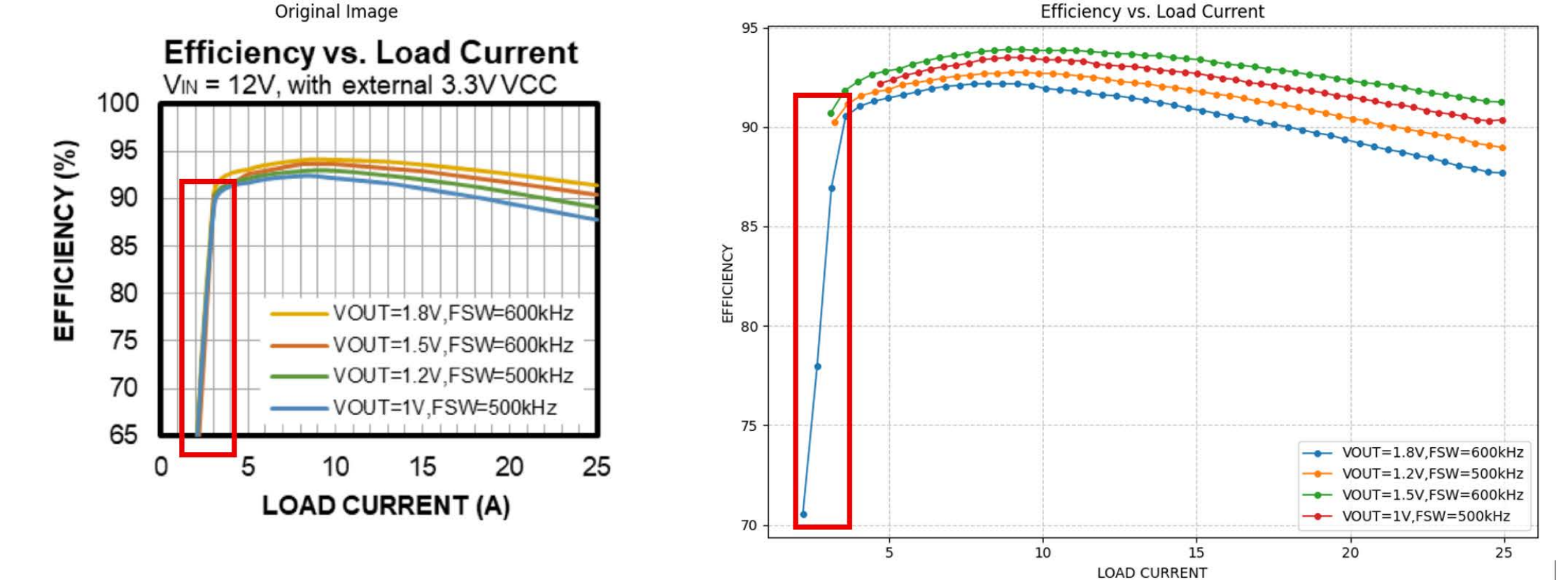}
        \caption{Sharp Overlaps}
        \label{fig:fail_overlap}
    \end{subfigure}\hfill
    \begin{subfigure}[b]{0.5\linewidth}
        \centering
        \includegraphics[width=\linewidth]{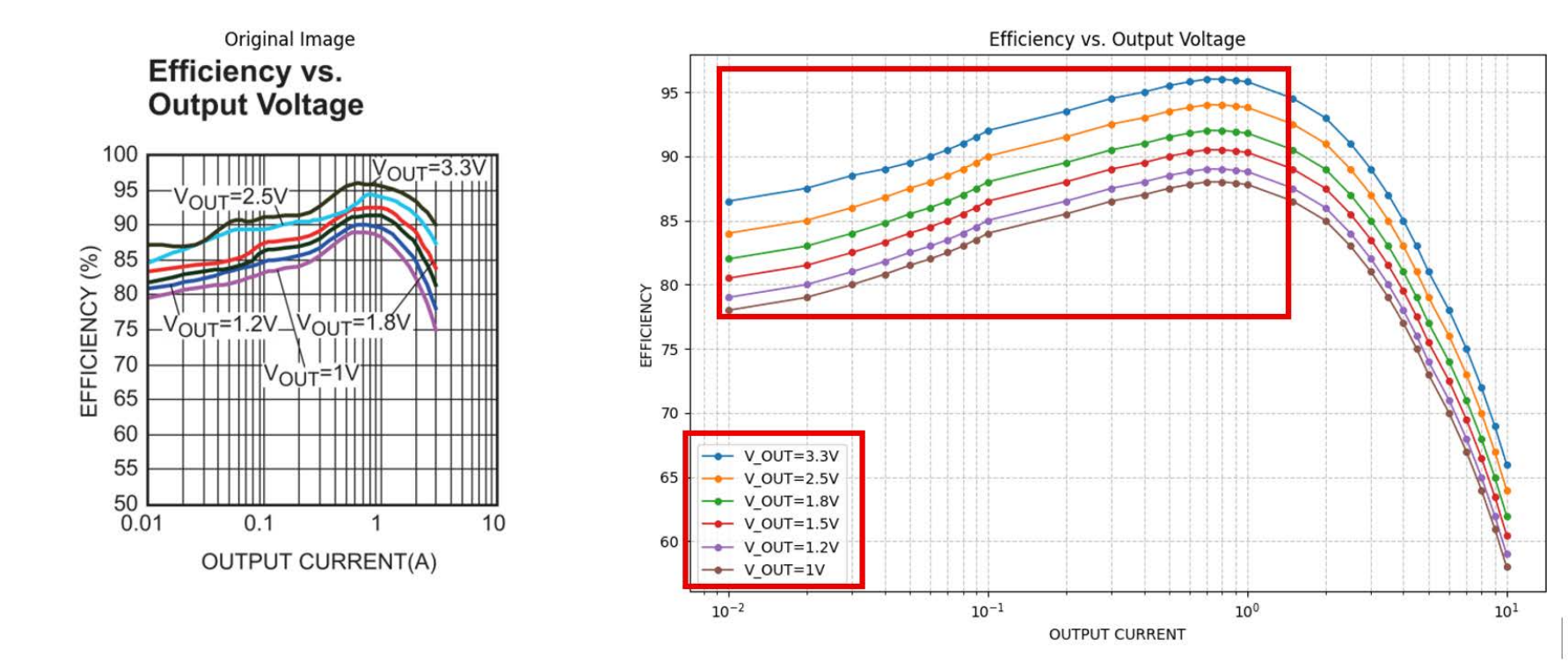}
        \caption{Clustered Curves}
        \label{fig:fail_cluster}
    \end{subfigure}

    \caption{Common failure modes of ChartZero. (a) Highly acute intersections can occasionally cause local instance swapping. (b) Densely clustered curve groups may lead to the detection of spurious "phantom" curves or merged segmentations.}
    \label{fig:failure_cases}
\end{figure}

Despite the robust zero-shot generalization of \textit{ChartZero}, we identify two primary failure modes where the current pipeline maintains structured limitations (Fig.~\ref{fig:failure_cases}).

Failure arises mainly in two settings: \textbf{sharp overlaps} (Fig.~\ref{fig:fail_overlap}), where acute intersections can cause local instance swaps, and \textbf{clustered curves} (Fig.~\ref{fig:fail_cluster}), where very small inter-curve spacing can produce merged or phantom detections. On the 1,000-chart validation split, these occur in approximately 4.2\% and 3.9\% of charts, respectively. In both cases, local geometric errors can propagate to downstream legend association, causing larger drops in ChartRM than IoU alone would suggest. These results indicate that improving long-range continuity and scale-adaptive supervision in dense regions remains an important next step.

\section{Conclusion}
To address the scarcity of annotated chart data, we propose ChartZero, a framework utilizing synthetic supervision and a novel GOI loss to achieve state-of-the-art generalization on in-the-wild plots. Beyond the model, we introduce the ChartRM metric to standardize end-to-end evaluation. Our benchmarking of current VLMs reveals a rapid evolution in chart parsing capabilities, suggesting a shift toward unified multimodal architectures for automated document digitization.


%
%
\bibliographystyle{splncs04}
\bibliography{main}
\end{document}